# Application of Unsupervised Outlier Detection to Improve Bird Audio Dataset Labeling

Bruce Collins
Independent Researcher
April 25, 2025

## Abstract

The Xeno-Canto bird audio repository is an invaluable resource for those interested in vocalizations and other sounds made by birds around the world. This is particularly the case for machine learning researchers attempting to improve on the bird species recognition accuracy of classification models. However, the task of extracting labeled datasets from the recordings found in this crowd-sourced repository faces several challenges. One challenge of particular significance to machine learning practitioners is that one bird species label is applied to each audio recording, but frequently other sounds are also captured including other bird species, other animal sounds, anthropogenic and other ambient sounds. These non-target bird species sounds can result in dataset labeling discrepancies referred to as *label noise*. In this work we present a cleaning process consisting of audio preprocessing followed by dimensionality reduction and unsupervised outlier detection (UOD) to reduce the label noise in a dataset derived from Xeno-Canto recordings. We investigate three neural network dimensionality reduction techniques: two flavors of convolutional autoencoders and variational deep embedding (VaDE (Jiang, 2017)). While both methods show some degree of effectiveness at detecting outliers for most bird species datasets, we found significant variation in the performance of the methods from one species to the next. We believe that the results of this investigation demonstrate that the application of our cleaning process can meaningfully reduce the label noise of bird species datasets derived from Xeno-Canto audio repository but results vary across species.

# 1  Introduction

Xeno-Canto.org provides access to a crowd-sourced repository of over 500,000 audio recordings of bird sounds along with the contributors' annotations and other metadata. This repository is an invaluable resource for anyone interested in real-world audio recordings of the variety of vocalizations and other sounds made by birds and other animals around the world. Many machine learning researchers use recordings from the Xeno-Canto repository as a source of training data for bird sound classification models and other machine learning projects. However, as with any unmoderated, crowd-sourced data there are no guarantees on the quality of the recordings or the validity of the metadata. Therefore, a machine learning researcher wishing to use this repository as a source of training data must screen and preprocess the data to ensure that it meets their needs.

This work describes a data processing pipeline that downloads audio recordings and associated metadata from Xeno-Canto.org, preprocesses the data to extract and screen samples and then attempts to identify samples that are not from the target bird species. This last step of detecting samples of sounds captured on a recording that were not produced by the recordist's target bird species (i.e., the labeled bird species) is particularly challenging. Discrepancies between the sounds captured in an audio sample and the bird species label that is applied to it is referred to as *label noise*. Even the most meticulous of recordists will often capture sounds other than the intended bird species sounds. While the Xeno-Canto platform provides a metadata field for such extraneous bird species this field cannot be relied on because it is not always filled out and may not be complete or correct. Even when other species are listed, there is no reliable way to associate these annotations to specific instances of sounds captured on the recording. In addition to other bird species vocalizations other animal sounds (e.g., frogs, insects, dogs), environment sounds (e.g., wind, rain, running water), human ambient sounds (e.g., cars, planes, trains) and even human voices, both intentional (e.g., audio annotations) and unintentional (e.g., children playing in the background) are sometimes captured on these audio recordings. While some of these errant sounds can easily be detected by a human screener, with the number of samples needed for a training dataset typically being large, employing such human screeners is very labor intensive and, in many cases, completely impractical. Alternatively, if all samples extracted from the audio recordings were included in the training dataset, the resulting label noise could be significant. The goal of the processing pipeline reported on here is to apply both standard signal processing techniques and unsupervised machine learning techniques to automate the cleaning process and to significantly reduce the label noise.

The main contributions of this work are the following:

1) Demonstration of how unsupervised outlier detection can be applied to real-world, bird sound datasets derived from Xeno-Canto with a non-trivial set of target species and the incumbent range of recording quality to detect outliers and reduce label noise.
2) Show the relative effectiveness of convolutional autoencoders, convolutional variational autoencoders and variational deep embedding architectures on this particular data cleaning task.

3) Provides insights into some of the parameters that influence the effectiveness of these unsupervised outlier detection techniques.

The software implementations for the various methods described in this article are available in (https://github.com/bcollins92078/bird_sounds_uod, n.d.)

## 2 Related Works

The topics of outlier detection and anomaly detection have been active areas of research for some time. A number of techniques and methodologies have been proposed to address specific applications (Wang, 2019) and problem formulations. Of relevance to this work are works on unsupervised outlier detection, cleaning and/or labeling of bird sound audio. (Denton, 2021) proposes combining an unsupervised sound separation model with a birdsong classifier as a method of improving birdsong classification performance. Similar to our work (Zhang, 2019) proposes training class-specific autoencoders but the proposed outlier detection is based on reconstruction error. However, when outliers exist in the training dataset, as is the case in many real-world applications, autoencoders tend to underestimate the reconstruction error of outliers and overestimate it for normal data (Abhaya, 2022).

The term *outlier* is used in this work to describe the sound samples extracted from an audio recording that were not produced by the bird species label which was applied to the recording. Some categories of outliers are quite dissimilar from the sounds that the target species produces. Some of the examples cited above such as non-bird animal sounds and anthropophonics (sounds generated from human activities) fall into this category. However, other outliers can be quite similar to a target bird sound such as sounds from other bird species. Outliers in the latter category can be very challenging to detect using unsupervised techniques. Another challenging scenario is when a bird species mimics the sounds of other birds. The classic example of this is the Northern Mockingbird but there are several others species that incorporate mimicry in their songs. Yet another challenging scenario is when the repertoire of the target bird species is so diverse that the representation learned by neural network appears to be less capable of differentiating outliers.

## 3 Dataset

Xeno-Canto is an online crowdsourced repository dedicated to freely sharing animal sound recordings from around the world (https://xeno-canto.org/, n.d.). Thousands of both amateur and professional recordists contribute their recordings to this repository which provides a vital resource for research scientist, birders and wildlife enthusiasts. The collection of bird sound recordings found in this repository contains over 857,000 recordings from more than 10,500 bird species as of October 2024.

The initial dataset used in this work consists of Xeno-Canto audio recordings and associated metadata downloaded for 50 bird species present in Southern California (Table 1). The number of recordings downloaded for each species and percentage of recordings which the metadata indicates contains songs, calls or both songs and calls samples or some other sound are shown for each of these 50 species. These latter descriptors were derived from the Xeno-Canto metadata *Type* field. Since this is a free-form field categories were assigned by simply searching for "song" and "call" strings and assigning the corresponding label if only one of these terms is found. The label *both* is assigned if both terms are found in the Type field and the label *other* is assigned if neither term is found (e.g., drumming for a woodpecker species). These simple categories provide an indication of the diversity and distribution of sounds captured in the dataset for each species. The number of recordings and the distribution and diversity of sounds captured for a bird species turn out to be important factors in the performance of dataset cleaning for that species.

The set of bird species in this work attempts to sample a meaningful range of outlier detection challenges. All the bird species included are land birds that might be heard in at least one of the many habitats in Southern California during at least some portion of the year. Most of these species are in the order Passeriformes (commonly referred to as songbirds) because these species make up more than half of all birds (https://en.wikipedia.org/wiki/List_of_bird_genera, n.d.) and they produce a huge diversity of sounds. However, other bird families are also included, such as raptors (Falconiformes), waterfowl (Anatidae) and hummingbirds (Trochilidae).

*Table 1: Bird species studied in order of English common name*

| Common name (English) | Species code | Genus | Species | No. files | % song | % call | % both | % other |
|---|---|---|---|---|---|---|---|---|
| Acorn Woodpecker | ACWO | Melanerpes | formicivorus | 196 | 8% | 86% | 4% | 3% |
| American Bushtit | BUSH | Psaltriparus | minimus | 162 | 7% | 88% | 4% | 1% |
| American Coot | AMCO | Fulica | americana | 231 | 3% | 95% | 1% | 1% |
| American Crow | AMCR | Corvus | brachyrhynchos | 234 | 2% | 95% | 1% | 1% |
| American Kestrel | MAKE | Falco | sparverius | 145 | 14% | 83% | 1% | 2% |
| Anna's Hummingbird | ANHU | Calypte | anna | 230 | 43% | 46% | 7% | 4% |
| Ash-throated Flycatcher | ATFL | Myiarchus | cinerascens | 208 | 20% | 69% | 11% | 0% |
| Audubon's Warbler (Yellow-rumped Warbler) | YRWA | Setophaga | auduboni | 133 | 49% | 35% | 15% | 2% |
| Bell's Vireo | BEVI | Vireo | bellii | 335 | 70% | 18% | 11% | 1% |
| Bewick's Wren | BEWR | Thryomanes | bewickii | 541 | 53% | 31% | 16% | 1% |
| Black Phoebe | BLPH | Sayornis | nigricans | 132 | 42% | 52% | 6% | 0% |
| Black-headed Grosbeak | BHGR | Pheucticus | melanocephalus | 224 | 70% | 20% | 10% | 0% |
| Blue-grey Gnatcatcher | BGGN | Polioptila | caerulea | 246 | 24% | 53% | 23% | 1% |
| California Quail | CAQU | Callipepla | californica | 179 | 18% | 56% | 25% | 1% |
| California Scrub Jay | CASJ | Aphelocoma | californica | 74 | 7% | 93% | 0% | 0% |
| California Thrasher | CATH | Toxostoma | redivivum | 117 | 72% | 23% | 4% | 1% |
| California Towhee | CATO | Melozone | crissalis | 92 | 33% | 64% | 3% | 0% |
| Cassin's Kingbird | CAKI | Tyrannus | vociferans | 127 | 19% | 78% | 3% | 0% |
| Cedar Waxwing | CEDW | Bombycilla | cedrorum | 96 | 5% | 86% | 8% | 0% |
| Common Poorwill | COPO | Phalaenoptilus | nuttallii | 176 | 48% | 43% | 9% | 1% |
| Common Yellowthroat | COYE | Geothlypis | trichas | 381 | 62% | 30% | 7% | 0% |
| Cooper's Hawk | COHA | Accipiter | cooperii | 184 | 4% | 92% | 1% | 4% |

| Common name (English) | Species code | Genus | Species | No. files | % song | % call | % both | % other |
|---|---|---|---|---|---|---|---|---|
| Eurasian Collared Dove | EUCD | Streptopelia | decaocto | 988 | 53% | 31% | 13% | 2% |
| Great Horned Owl | GHOW | Bubo | virginianus | 504 | 43% | 42% | 7% | 8% |
| Greater Roadrunner | GRRO | Geococcyx | californianus | 138 | 42% | 29% | 3% | 26% |
| House Finch | HOFI | Haemorhous | mexicanus | 282 | 56% | 25% | 19% | 0% |
| House Wren | HOWR | Troglodytes | aedon | 1196 | 69% | 22% | 7% | 2% |
| Killdeer | KILL | Charadrius | vociferus | 293 | 3% | 92% | 3% | 2% |
| Lazuli Bunting | LAZB | Passerina | amoena | 109 | 89% | 6% | 5% | 0% |
| Lesser Goldfinch | LEGO | Spinus | psaltria | 230 | 38% | 43% | 18% | 0% |
| Mountain Chickadee | MOCH | Poecile | gambeli | 225 | 34% | 55% | 11% | 0% |
| Mourning Dove | MODO | Zenaida | macroura | 197 | 75% | 11% | 3% | 11% |
| Northern Flicker | NOFL | Colaptes | auratus | 207 | 21% | 68% | 5% | 5% |
| Nuttall's Woodpecker | NUWO | Dryobates | nuttallii | 51 | 2% | 67% | 2% | 29% |
| Oak Titmouse | OATI | Baeolophus | inornatus | 119 | 44% | 39% | 17% | 0% |
| Orange-crowned Warbler | OCWA | Leiothlypis | celata | 240 | 68% | 27% | 5% | 0% |
| Pacific-slope Flycatcher | PSFL | Empidonax | difficilis | 193 | 26% | 56% | 3% | 15% |
| Phainopepla | PHAI | Phainopepla | nitens | 74 | 9% | 68% | 23% | 0% |
| Red-breasted Sapsucker | RBSA | Sphyrapicus | ruber | 39 | 5% | 62% | 0% | 33% |
| Red-shouldered Hawk | RSHA | Buteo | lineatus | 141 | 4% | 94% | 0% | 2% |
| Red-winged Blackbird | RWBL | Agelaius | phoeniceus | 564 | 20% | 41% | 38% | 1% |
| Rock Wren | ROWR | Salpinctes | obsoletus | 108 | 59% | 37% | 4% | 0% |
| Say's Phoebe | SAPH | Sayornis | saya | 69 | 55% | 41% | 4% | 0% |
| Scaly-breasted Munia | SBMU | Lonchura | punctulata | 141 | 4% | 89% | 6% | 2% |
| Song Sparrow | SOSP | Melospiza | melodia | 777 | 71% | 22% | 5% | 1% |
| Spotted Towhee | SPTO | Pipilo | maculatus | 588 | 65% | 29% | 6% | 0% |
| Western Meadowlark | WEME | Sturnella | neglecta | 306 | 63% | 15% | 22% | 1% |
| White-breasted Nuthatch | WBNU | Sitta | carolinensis | 335 | 20% | 68% | 11% | 1% |
| White-crowned Sparrow | WCSP | Zonotrichia | leucophrys | 353 | 63% | 22% | 15% | 0% |
| Yellow-breasted Chat | YBCH | Icteria | virens | 460 | 75% | 17% | 7% | 1% |

# 5  Audio Pre-processing

The audio recordings in the Xeno-Canto bird sounds repository are stored in MP3 format which is a digital audio format ubiquitously used to store good quality but compressed audio. These MP3 files are downloaded using the API described at (https://xeno-canto.org/, n.d.). Files are read into Python code using the Librosa library which returns a floating-point time series resampled at the default sampling rate of 22.05 kHz. Each time series is then segmented based on a silence detection algorithm very similar to pydub.silence.detect_silence and the non-silence sections are converted into mel-spectrograms. Selection of the spectrogram resolution is an important decision. Values that are too high result in issues referred to as *the curse of dimensionality*. Selecting values too low result in loss of too much information. We settled on a resolution of 32 Mel frequency bands by 40 timeframes as a compromise.

Care was taken in the segmentation algorithm to avoid ignoring short segments that might contain bird "chip-calls" which can be very short in some species. During this segmentation process a signal-to-interference and noise ratio (SINR) is calculated for each segment relative to the average energy level in the silence segments. This SINR value is used in a downstream screening process.

The above segmentation and conversion to mel-spectrograms yields a 2-dimensional, monochrome image for each non-silence segment. The next step in the preprocessing is to screen these spectrograms for SINR. The minimum SINR of the spectrograms to be used in downstream processing is a critical hyperparameter. Setting this value too low hampers the ability of the process to cluster target species vocalizations and to detect outliers. However, increasing this minimum SINR reduces the number of spectrograms retained and thus reduces the size of the dataset. A second SINR screening criterion is applied to remove background sounds. In most recordings the target sounds are among the highest energy level sounds on the recording. Therefore, it should be possible to retain these target sounds and remove some of the unintentional ones by removing segments with an SINR below some minimum threshold relative to the *highest level* sounds on the recording. The 75$^{th}$ percentile of the segment SINRs on a recording was determined to be a good measure of the highest level sounds and a minimum threshold of 5dB below that was used by default to remove background sounds.

After screening for SINR a fixed time duration portion is then extracted from each of these spectrograms (referred to as *clips* hereafter) to be used during the downstream data cleaning process steps. Since this cleaning process is executed on a per-species basis the duration of these clips is selected based on the vocalization characteristics of the species but generally ~1 second duration was found to work well for most species. If the duration of a particular spectrogram is shorter than this fixed clip duration selected for the species, then padding is applied. An important consideration during this process is time alignment of the clips. It turns out that time shifts in spectrogram have a significant impact on the downstream dimensionality reduction step. Therefore, the feature extraction and padding are performed to place the center-of-energy of the clip at the center of the clip timespan. Figure 1 shows a high-level summary of the audio preprocessing steps.

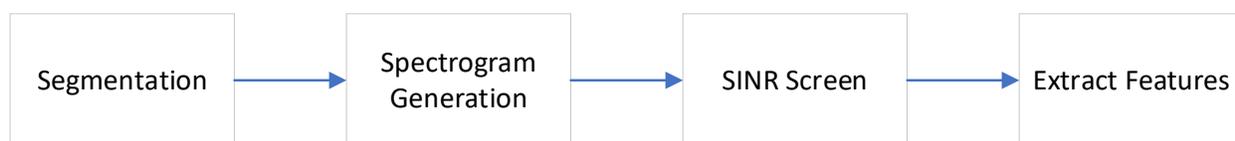

*Figure 1. Audio preprocessing block diagram*

# 6   Dimensionality Reduction

Any reasonable resolution spectrogram will result in high dimensional data. We experimented with resolutions as low as 16 frequencies X 20 timeframes and while we found this resolution provides only marginally recognizable visualizations the resulting 320 dimensions rendered all attempts to detect more than the most egregious outliers ineffective. We found that clustering attempts on datasets for bird species that have two quite distinct vocalization types failed to cleanly separate examples into different clusters. And while we did not find that clustering effectiveness to be synonyms with effective outlier detection the ability to distinguish between significantly different vocalizations does indicate that outlier sounds might also be separable. We attribute this failure to separate distinct target species vocalizations to the "curse of dimensionality" and investigated dimensionality reduction techniques.

Dimension reduction is commonly defined as the process of mapping high-dimensional data to a lower-dimensional embedding  (Engel, Hüttenberger, & Hamann, 2012). Dimensionality reduction techniques have been a topic of active research for a couple of decades (Engel, Hüttenberger, & Hamann, 2012). In recent years artificial neural network (ANN) based approaches have emerged that automatically learn highly nonlinear functions that perform the mapping to lower dimensional *latent spaces* while preserving the most significant features of the original data. Referred to as representation learning or feature extraction these neural networks learn the features in the data based on their training objective function. Since the objective here is to improve the labels in a bird species dataset we cannot apply a supervised neural network training methodology. Here we apply only unsupervised methods.

One popular unsupervised methodology involves training an encoder-decoder type neural network referred to as an autoencoder. In fact, autoencoders are the core of all unsupervised deep outlier detection models (Chalapathy, 2019). The encoder portion of this ANN architecture is designed to map its input, high-dimensional data into a lower dimensional latent space. The decoder portion takes the latent space representation as input and attempts to reconstruct the original high-dimensional data that was input. The autoencoder's encoder and decoder parts are trained jointly by adjusting the neural network parameters of each to minimize the difference between the input and reconstructed output (the *reconstruction loss*). While the objective function of the simplest autoencoder (referred to here as the *conventional* autoencoder) only contains a reconstruction loss term, many autoencoder variants have been proposed over the past decade or more that introduce additional terms.

The variational autoencoder (VAE) is one such variant of the conventional autoencoder which some have claimed to extract features from high-dimensional image datasets more stably and accurately (Dong, 2018). While the VAE neural network architecture is similar to the conventional autoencoder, consisting of an encoder and a decoder, the encoder generates a mean and variance vector as latent space representations of each high-dimension input. And in addition to the autoencoder's reconstruction loss term the VAE loss function includes a term that encourages the learned latent space representation to follow a multivariate, unimodal Gaussian distribution (Doersch, 2016). These changes result in the VAE model's generative capabilities and its continuous latent space representation that attempts to preserve the most important features that explain most of the variance of the input data. Note that in our application the VAE's generative capability is not essential. Here we are only concerned with outlier detection.

Since we are processing bird sound recordings as spectrograms which are two dimensional images we chose to employ convolutional layers in both the conventional and variational autoencoder architectures. Convolutional neural networks (CNNs) have become dominant in image processing tasks such as image classification, object detection, and segmentation due to their superior performance at extracting features from images. When convolutional layers are included in an autoencoder architecture it is referred to as a convolutional autoencoder (CAE). Figure 1 shows a high-level architecture of the convolutional encoder-decoder network used in this study for the CAE *and* CVAE implementations and the yet to be discussed VaDE implementation. It consists of three convolutional layers in the encoder followed by a fully connected layer and a fully connected layer followed by three deconvolutional layers (ConvT) in the decoder. The details of the Latent Space Distribution Model are different for each of the implementations.

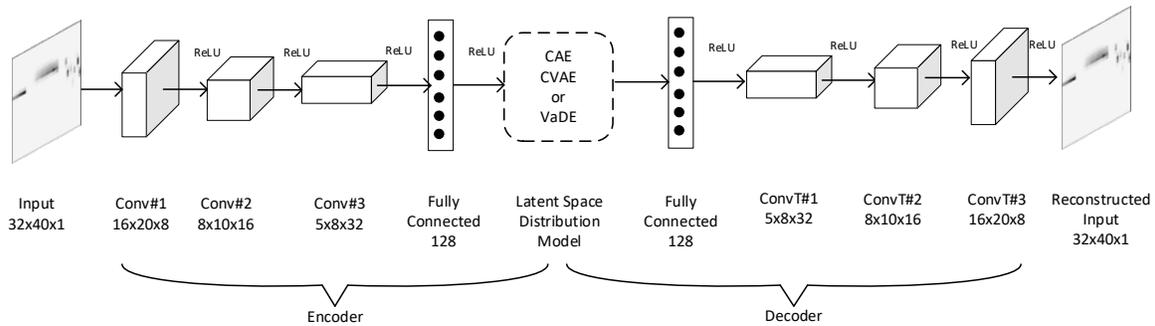

*Figure 2. Convolutional neural network autoencoder architecture*

During training the encoder of the conventional CAE learns a function that maps each high-dimensional image in the dataset to a point in the lower dimensional latent space and the decoder learns a function that maps each point latent space point to an approximate reconstruction of the corresponding input image. So, the latent space distribution model in this case is not probabilistic at all but direct mappings.

The CVAE encoder, on the other hand, maps high-dimensional images to a probabilistic (unimodal Gaussian) distribution in the latent space, rather than specific points. It maps each input image to a mean vector and variance vector of the dimension of the latent space and the decoder samples this distribution and maps that sample back to a high-dimensional image.

# 7 Unsupervised Outlier Detection

After dimensionality reduction any of the popular clustering techniques might be applied to obtain some separation of the various vocalization types that might be represented in the data. However, while such separation might provide some indication that a useful representation was learned during dimensionality reduction the goal here is not clustering. Instead, the goal here is to detect anomalous clips (i.e., outliers). For the application of a clustering algorithm to be useful in identifying outliers, anomalous data points must be mapped into regions of the latent space removed from where concentrations of target bird species produced clips. Figure 3 shows a scatter plot of the two-dimensional latent space of a CAE trained on the pre-processed samples from Yellow-rumped Warbler. The colors indicate the descriptor attached to the file from which the clip was extracted (as described in the Dataset section above). This species has distinct song and call vocalizations and this is apparent in the scatter plot. Notice that there is one cluster in the upper left where mostly inputs labeled song are mapped and actually two clusters in the lower right where mostly call labeled inputs are mapped. The fact that there are two call clusters indicates that this CAE instance has learned a representation that distinguishes two types of call vocalizations. In addition, also notice that several points in the scatter plot appear at various distances from either of the song or call clusters. These points are candidates for outlier designations. We want to apply an algorithm that will identify the major clusters so that the data points that do not fall near any of these clusters can be algorithmically detected.

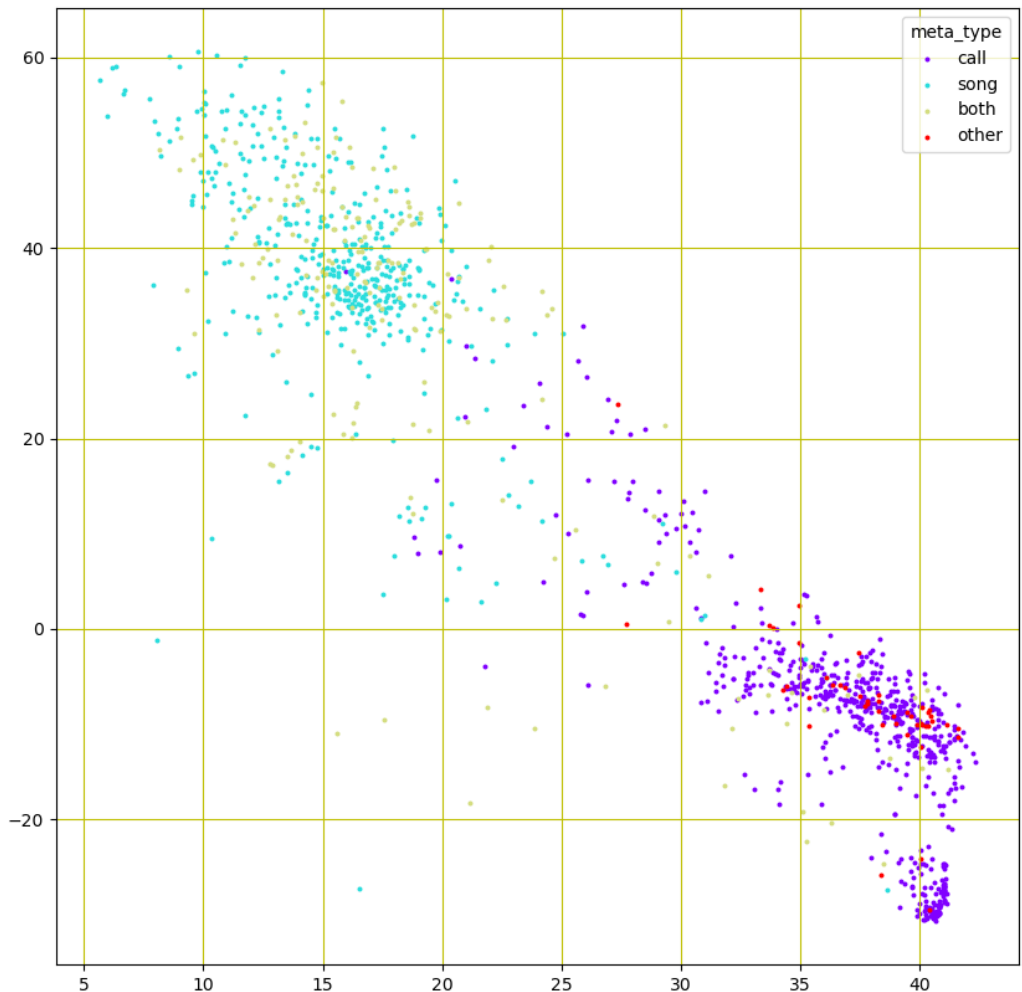

*Figure 3. Scatter plot of CAE latent space for Yellow-rumped Warbler dataset*

An issue with most of the popular clustering algorithms is the need to specify the number of clusters (alone with other hyperparameters) a priori. This is difficult in our case since for some bird species the different vocalizations type can be quite similar to other types from that same species. So the autoencoder may not learn a representation that distinguishes between them and these similar sounds could end up in the same cluster. Furthermore, one cannot predict which of the sounds uttered by a bird species have been captured in sufficient numbers in the set of recordings available on Xeno-Canto to allow for them to constitute a cluster. One approach to overcoming such hyperparameter issues is to perform a grid search but given the number of species involved in this study and the manual effort required to assess the search results, this was deemed impractical.

The hierarchical agglomerative clustering (HAC) algorithm is one well known clustering method that avoids the need to specify the number of clusters (Nielsen, 2016). HAC builds a binary merge tree, starting from individual data elements at the leaves (singleton sets) and proceeding by merging the "closest" two subsets until the root of the tree that contains all the elements is reached. The algorithm provides flexibility in the distance metric used and the "linkage" between two subsets. The distance metric provides a measure of the distance between any two data points. The most common distance metric is *Euclidean distance*; however, several other options and even custom distance metrics are commonly supported by library implementations (e.g., SciPy and Scikit-learn). The term linkage is used in this context to describe how the distance between two subsets of data points is calculated based on the distance metric selected. The standard implementation defines four linkage methods – single linkage, complete linkage, average linkage and Ward linkage (Nielsen, 2016). The average linkage defines the distance between two subsets as the average of the pairwise distances between each pair of points in the two subsets. Average linkage was used in this study because of its intuitive characteristics and its monotonicity properties.

Figure 4 shows the truncated binary tree diagram for the hierarchical agglomeration clustering of the CAE latent space for Yellow-rumped Warbler shown in Figure 3. This tree diagram is referred to as a *dendrogram* and it details the HAC execution from the beginning when each data point constitute its own cluster (i.e., singletons) until they have all merged into one cluster at the root of the tree. For such a diagram to be visually usable this dendrogram is truncated to start at the last 12 flat clusters and shows the final 11 merges into the root of the tree. The sizes of clusters are shown in parentheses along the horizontal axis and the merge distances on the vertical axis. It provides a visual representation of how HAC evolves over the final 11 steps of the clustering process. Notice that by this point in the HAC process two large clusters have already formed, clusters #5 and #1 (clusters are number from 1 to 12 from left to right on the horizontal axis). These clusters contain 33.3% and 24.0%, respectively, of all the samples and contain examples of two of this species' two most common vocalization types – a short duration *call* and a longer duration *song*. Also note that clusters #1, #2 and #3 merge in short order indicating that the latent space representation of these samples are close to each other and, if the autoencoder did its job correctly, then all the samples in these clusters are in some sense similar. Likewise, clusters #4, #5, #6 and #7 also merge early in these final steps of agglomeration indicating that these samples are similar to each other. Even though clusters #6 and #7 are quite small, the fact that they merge with big cluster #5 early on indicates that they are less *anomalous* than other clusters that merge later. Clusters #8 thru #12 merge with each other first and then merge with the group of clusters formed by #4 thru #7. Finally, the group of clusters formed by #1 thru #3 merge at the root. The fact that the two biggest clusters among the 12 initial clusters don't merge until the root node of the tree and only after a big distance jump indicates that their latent space representations are significantly different.

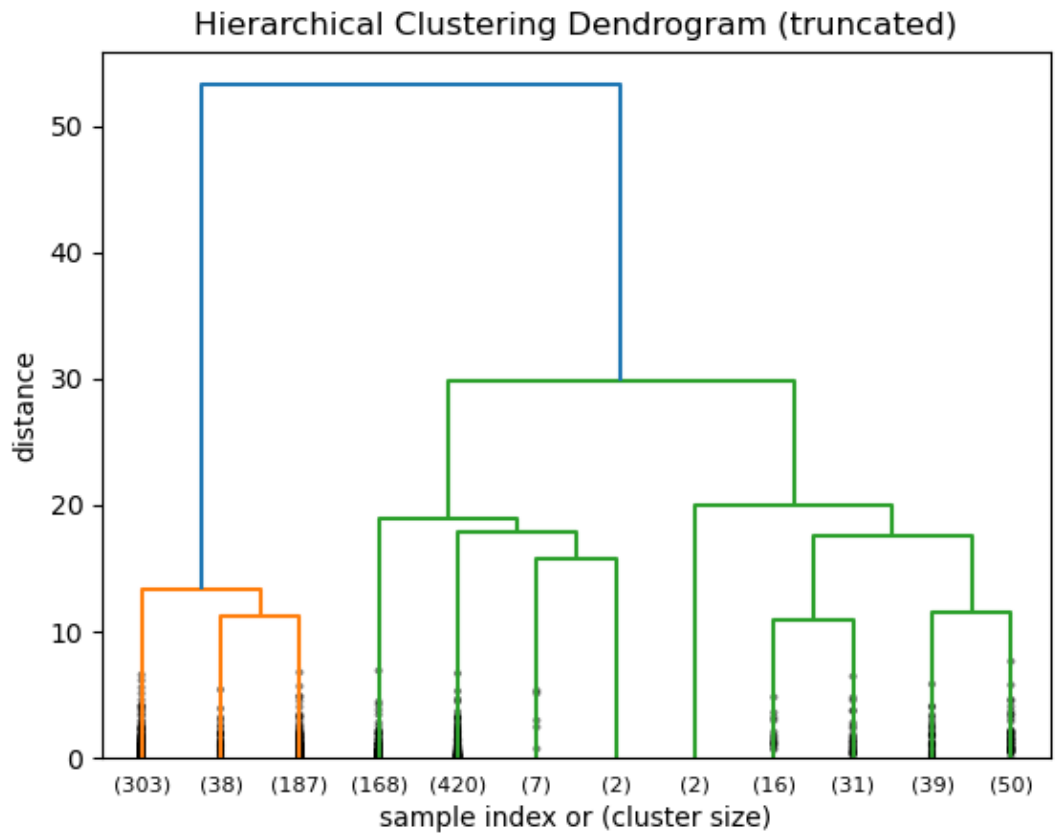

*Figure 4: Example truncated dendrogram of 2-D latent space learned by a CAE model for Yellow-rumped Warbler*

Figure 5 shows a scatter plot of the same CAE latent space for Yellow-rumped Warbler shown in Figure 3 but this time the colors indicate the HAC cluster assignments shown in the dendrogram of Figure 3. Note that the two large clusters, #5 and #1 are clearly represented as part of the large song and call collections of points noted in previously. Cluster #4 contains the call samples mentioned earlier that are in some sense similar to, but distinct from the bulk of the call samples in cluster #5. The candidate outlier clusters mentioned above, #9, #10, #11 and #12, occupy locations on the periphery and are much more scattered. Also note that clusters #6 and #7 are also quite small and scattered so might also be a candidate for outlier designation. However, they are closer to big cluster #5 in this encoding and therefore are considered less anonymous.

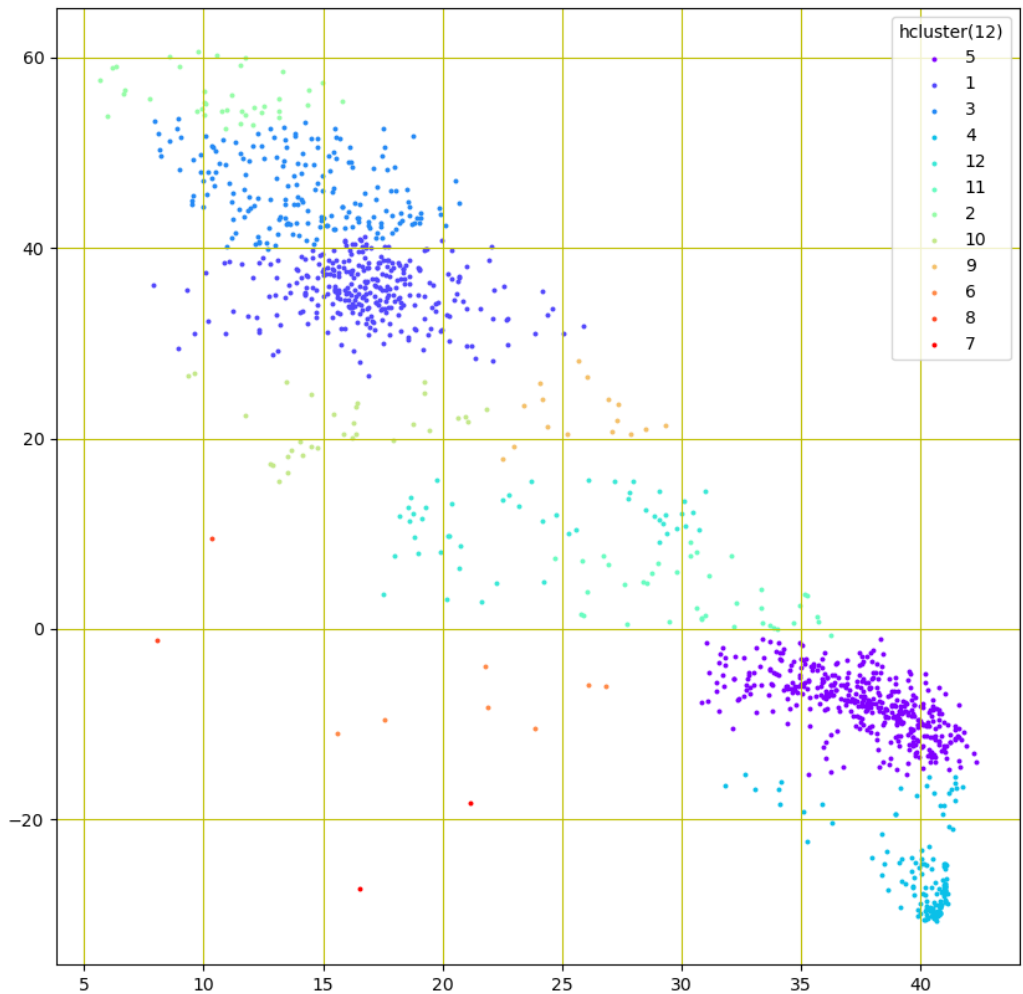

*Figure 5. Example scatter plot of 2-D latent space learned by a CAE model for Yellow-rumped Warbler*

Figure 4 and Figure 5 illustrate the process by which outlier detection can be performed using an autoencoder trained on a one-species dataset in order to reduce dimensionality and then application of HAC to cluster the latent space representations of the data points and delineate smaller, late merging clusters that might consist of outliers. However, these figures are for illustration only since extracting only 12 flat clusters did not provide enough resolution to reliably discern the outliers before they merge in our experiments. In addition, a two-dimensional latent space is not sufficient for most bird species datasets. We have found that 50 flat clusters and 10 latent dimensions as defaults provided much better performance.

We observed significant variability in the latent space representations learned from one autoencoder model instance to the next. While most instances exhibited similar clustering characteristics and detected the most egregious outliers, some samples are flagged as outliers by some model instances but not by others. We found that taking a simple ensemble approach of training multiple models on the same single-species input data, applying HAC to identify candidate outliers from each model and then designating as an outlier any clip that a *majority* of the models flagged as an outlier improved performance significantly. Majority voting is the most popular and intuitive ensemble combination method (Sun, 2022).

Our UOD method can now be concisely described:

*Table 2. UOD method #1*

| |
|---|
| Hyperparameters:<br>• $N$ – number of models to train<br>• $C$ - number of flat clusters<br>• $D$ – maximum number of samples to be discarded as outliers<br>• $B$ – minimum cluster size to be considered a big cluster |
| Algorithm:<br>1. For $N$ autoencoder models<br>   a. Train model on single-species, preprocessed audio spectrogram dataset<br>   b. Apply HAC models learned latent space representation of the input data points<br>   c. Extract $C$ flat clusters from the hierarchical clustering<br>   d. Designate any cluster that contains at least *B%* of total dataset size as a big clusters<br>   e. Compute the distances between each cluster and each of the big clusters (these distances are computed as a byproduct of the HAC process)<br>   f. For each cluster take the minimum distance to any big cluster as its $d_{big}$<br>   g. Starting with the cluster with largest $d_{big}$ and proceeding in order of decreasing $d_{big}$ increasing cluster size, designate the members of the cluster to be candidate outliers until the total number of such candidates reaches, but does not exceed, $D$<br>2. Tally the number of candidates outlier designations for each data point and designate any points with a majority tally as an outlier |

# 8  Variational Deep Embedding

While the autoencoder models provide significant dimensionality reduction allowing the concatenated HAC to effectively detect outliers in the bird audio datasets, we wondered whether an approach that explicitly attempts to model the distinct sounds that may be represented in the single-species input dataset might be able to detect more outliers. Referring again to the scatter plot in Figure 3 where there appeared to be 3 clusters, one for data points from recordings with a song descriptor and two clusters of points from files with call descriptor. Our thought was that if we could explicitly model 3 clusters in such a scenario maybe outliers that were more similar to these call clusters might be exposed. There have been a number of proposals for extending the VAE concept to use a Gaussian Mixture Model (GMM) prior distribution instead of the unimodal Gaussian with the explicit motivation of supporting the clustering task. One of the earliest was Variational Deep Embedding (VaDE) (Jiang, 2017). VaDE was originally proposed as a generative clustering method that combines VAE with Gaussian Mixture Models. It simultaneously learns feature representations and cluster assignments using deep neural networks.

VaDE models the generative process of the data as a GMM rather than the single Gaussian prior used in VAE models. In theory each of the distinct sound types captured in the dataset for a species in sufficient numbers can be assigned to a separate cluster and clips that are not well represented in the dataset (e.g., the outliers) should be distinguishable based on the probability of cluster membership. This probability of cluster membership is a byproduct of GMM and provides a natural mechanism for scoring anomalousness.

A brief overview of the steps in the Variational Deep Embedding (VaDE) algorithm are as follows:

1) Use the CVAE encoder to map the input data into a latent space.
2) Model the latent space using a GMM to capture the clustering structure.
3) Perform variational inference to optimize the parameters of both the CVAE and the GMM.
4) Assign data points to clusters based on the learned GMM in the latent space.

For a more detailed explanation of VaDE refer to (Jiang, 2017).

In order to avoid issues with a weak reconstruction term if simple random initialization of network parameters, the VaDE algorithm calls for *pretraining* the network using a conventional autoencoder before training with the full VaDE latent space distribution model. The authors claimed that only a few epochs of pretraining are required for this initialization; however, we used this pretraining step to train our CAE models.

*Table 3. UOD method #2*

| Hyperparameters: |
|---|
| • $N$ – number of models to train |
| • $C$ - number of clusters |
| • $D$ – maximum number of samples to be discarded as outliers |

| Algorithm: |
|---|
| 3. For $N$ VaDE models |
|     a. Train model on single-species, preprocessed spectrogram dataset |
|     b. Compute probability of each latent space data point given each GMM cluster (these probabilities are computed as a byproduct of GMM) |
|     c. Starting with lowest probability of cluster membership, designate data points to be candidate outliers until the total number of such candidates reaches $D$ |

> 4. Tally the number of candidates outlier designations for each data point and designate any points with a majority tally as an outlier

# 9 Assessing Performance of UOD.

A fundamental challenge that exists when applying UOD to real, unlabeled data is assessing the effectiveness of the resulting outlier detection. Unsupervised model selection using only the input data and without ground-truth labels is a notoriously difficult and unsolved problem (Ma, 2023). In this work we chose to sample the outlier classes of each of the model ensembles to estimate the true positive rate (TPR). By focusing only on the outlier classes and the fact that the number of outliers detected is constrained by a hyperparameter the sampling task is rendered more tractable. TPR provides an appropriate performance metric for our application because the goal is to detect as many of the non-target bird specs sounds as possible within the constraint imposed by the maximum discards hyperparameter. However, comparing TPR of different ensembles is only fair if the size of the outlier class of each is roughly the same. This is clear when one considers the scenario where one ensemble flags only one segment as anomalous and that one is indeed an outlier. The TPR for that ensemble would be a perfect 1.0 while another ensemble that flags more segments but with some number of false positives would not compare favorably. Therefore, the number of models of an ensemble required to agree to detect an outlier is adjusted so that the outlier class size of the ensembles to be compared are all as close as possible.

Of course, such an assessment does not provide any information about the false negative rate which is of interest but was deemed to be too expensive in terms of human effort to sample the inlier class for all UOD model ensembles. However, this *was* done in a few case as will be discussed later.

# 10 Sampling Methodology

The methodology used to sample the clips flagged by an ensemble is described in this section. The sampling was performed by the author who is a birding enthusiast but not an expert in identifying birds by ear. Therefore, the sampling workflow began with a review the sound recordings from the species on the companion website to The Peterson Field Guide to Bird Sounds (Pieplow, 2019). Then software was executed for the species that randomly selects from the clips flagged by the ensemble as outliers and displays the spectrogram of the clip and plays the audio of the entire segment that the clip was extracted from. The software then pauses for input from the user that consists of an indication that the segment is an outlier or that it is an acceptable species sound (an inlier) or that the user could not determine either way (indeterminant). Any other value entered simply replays the audio. Note that there was ample opportunity for judgement in practice, so the following are some of the rules used in deciding.

- If multiple sounds occur in the segment, then the decision is based on whether the non-species sound is strongly visible in the spectrogram of the clip
- If the sound is unusual but reasonably attributable to the target species, then the decision is that the clip is an inlier

In addition to the above input prompt, the software also accepts an optional text string as a comment. This was useful for identifying false positive tendencies for an ensemble.

# 11 Results

Table 4 shows the estimated TPR for each of the model ensembles applied to each bird species based on our sampling of the outlier class of each. Estimates were computed for a confidence interval of 95% and the margin of error (MoE) shown in the table. The number of clips, species sound diversity, entropy and cause factor columns will be discussed later in this section.

*Table 4. Best TPR performance of each ensemble type for each species included in the study*

| species code | No. clips | diversity | entropy | CVAE | MoE | CAE | MoE | VaDE | MoE | best | cause factor |
|---|---|---|---|---|---|---|---|---|---|---|---|
| ACWO | 2151 | medium | 6.7963 | 0.602 | 5.9% | 0.529 | 3.2% | 0.352 | 3.1% | 0.602 | |
| BUSH | 1510 | low | 4.5636 | 0.648 | 4.0% | 0.546 | 4.3% | 0.517 | 3.4% | 0.648 | |
| AMCO | 1438 | low | 6.4874 | 0.79 | 2.1% | 0.659 | 2.8% | 0.238 | 3.7% | 0.79 | |
| AMCR | 1595 | medium | 6.0612 | 0.704 | 4.4% | 0.773 | 3.0% | 0.304 | 3.6% | 0.773 | |
| AMKE | 574 | medium | 6.0018 | 0.429 | 2.8% | 0.777 | 0.0% | 0.485 | 6.7% | 0.777 | |
| ANHU | 1043 | low | 5.7423 | 0.886 | 4.9% | 0.892 | 2.3% | 0.588 | 4.9% | 0.892 | |
| ATFL | 4070 | low | 3.3521 | 0.925 | 3.8% | 0.935 | 4.8% | 0.889 | 6.4% | 0.935 | |
| YRWA | 1262 | medium | 4.4368 | 0.709 | 5.1% | 0.787 | 6.7% | 0.642 | 5.7% | 0.787 | |
| BEVI | 3223 | medium | 6.2013 | 0.522 | 4.0% | 0.398 | 4.3% | 0.261 | 3.6% | 0.522 | 3 |
| BEWR | 5591 | very high | 5.9190 | 0.148 | 4.5% | 0.19 | 2.5% | 0.179 | 2.5% | 0.19 | 3 |
| BHGR | 1840 | medium | 4.7980 | 0.79 | 4.0% | 0.752 | 3.0% | 0.73 | 3.7% | 0.79 | |
| BLPH | 2612 | low | 3.9297 | 0.325 | 4.1% | 0.333 | 3.7% | 0.365 | 3.6% | 0.365 | 4(0.0%) |
| BGGN | 2678 | high | 5.8110 | 0.284 | 3.4% | 0.309 | 3.7% | 0.31 | 3.5% | 0.31 | 3 |
| CAQU | 1769 | medium | 5.0482 | 0.221 | 4.9% | 0.347 | 4.9% | 0.495 | 5.0% | 0.495 | |
| CASJ | 911 | medium | 6.4874 | 0.361 | 3.6% | 0.448 | 5.3% | 0.529 | 4.5% | 0.529 | |
| CATH | 1380 | very high | 4.9678 | 0.287 | 4.4% | 0.357 | 4.0% | 0.357 | 4.6% | 0.357 | |
| CATO | 1177 | low | 2.7885 | 0.17 | 3.4% | 0.163 | 2.8% | 0.314 | 1.9% | 0.314 | 4(0.8%) |
| CAKI | 1785 | medium | 5.3328 | 0.1 | 2.7% | 0.437 | 5.2% | 0.286 | 3.1% | 0.437 | |
| CEDW | 743 | low | 5.0847 | 0.879 | 2.6% | 0.652 | 8.0% | 0.617 | 0.0% | 0.879 | |
| COPO | 2553 | low | 4.5177 | 0.531 | 5.0% | 0.433 | 3.4% | 0.379 | 3.5% | 0.531 | |
| COYE | 3203 | medium | 4.9883 | 0.704 | 4.9% | 0.57 | 3.7% | 0.592 | 3.6% | 0.704 | |
| COHA | 1113 | low | 5.9682 | 0.63 | 3.6% | 0.7 | 3.1% | 0.694 | 3.9% | 0.7 | |

| species code | No. clips | diversity | entropy | CVAE | MoE | CAE | MoE | VaDE | MoE | best | cause factor |
|---|---|---|---|---|---|---|---|---|---|---|---|
| EUCD | 6065 | low | 5.5768 | 0.783 | 5.3% | 0.634 | 5.5% | 0.479 | 5.1% | 0.783 | |
| GRRO | 2045 | low | 4.8877 | 0.632 | 4.0% | 0.571 | 6.9% | 0.585 | 6.9% | 0.632 | |
| GHOW | 1693 | medium | 6.6381 | 0.843 | 4.6% | 0.694 | 4.2% | 0.414 | 4.2% | 0.843 | |
| HOFI | 2894 | high | 4.9956 | 0.522 | 3.6% | 0.466 | 3.9% | 0.413 | 3.5% | 0.522 | |
| HOWR | 4755 | high | 6.2934 | 0.2973 | #### | 0.3616 | 3.00% | 0.2812 | 3.00% | 0.3616 | 3 |
| KILL | 1899 | medium | 4.6039 | 0.659 | 4.4% | 0.605 | 2.7% | 0.433 | 3.3% | 0.659 | |
| LAZB | 687 | medium | 5.6260 | 0.755 | 3.1% | 0.896 | 3.2% | 0.571 | 4.6% | 0.896 | |
| LEGO | 2323 | high | 5.1757 | 0.51 | 3.6% | 0.62 | 4.7% | 0.344 | 4.4% | 0.62 | |
| MOCH | 1936 | medium | 5.8170 | 0.429 | 3.1% | 0.39 | 4.1% | 0.36 | 5.8% | 0.429 | 3 |
| MODO | 1351 | low | 5.3124 | 0.936 | 2.2% | 0.857 | 5.6% | 0.845 | 2.8% | 0.936 | |
| NOFL | 1457 | medium | 5.1275 | 0.544 | 4.9% | 0.55 | 4.4% | 0.735 | 3.8% | 0.735 | |
| NUWO | 470 | medium | 4.9711 | 0.472 | 0.0% | 0.3 | 2.3% | 0.659 | 4.4% | 0.659 | |
| OATI | 1496 | high | 5.3320 | 0.382 | 3.4% | 0.3 | 4.5% | 0.34 | 2.6% | 0.382 | |
| OCWA | 1730 | low | 5.0829 | 1 | 5.1% | 0.787 | 3.9% | 0.785 | 4.1% | 1 | |
| PSFL | 2738 | low | 3.9113 | 0.962 | 3.7% | 0.934 | 1.5% | 0.74 | 3.8% | 0.962 | |
| PHAI | 1284 | medium | 4.2889 | 0.578 | 5.3% | 0.505 | 4.1% | 0.316 | 2.1% | 0.578 | |
| RBSA | 226 | medium | 5.8594 | 0.333 | | 0.25 | | 0.273 | | 0.333 | 2,3 |
| RSHA | 639 | medium | 5.7184 | 0.547 | 2.6% | 0.482 | 3.1% | 0.583 | 3.9% | 0.583 | |
| RWBL | 7003 | very high | 4.7980 | 0.206 | 3.9% | 0.362 | 4.4% | 0.302 | 5.5% | 0.362 | |
| ROWR | 1806 | very high | 5.5969 | 0.109 | 5.0% | 0.174 | 3.7% | 0.258 | 2.6% | 0.258 | 3 |
| SAPH | 1199 | low | 4.7753 | 0.691 | 4.0% | 0.634 | 2.8% | 0.588 | 2.9% | 0.691 | |
| SBMU | 565 | low | 4.2428 | 0.756 | 6.5% | 0.636 | 6.5% | 0.755 | 2.9% | 0.756 | |
| SOSP | 6797 | high | 4.7882 | 0.425 | 3.2% | 0.46 | 4.2% | 0.478 | 5.2% | 0.478 | |
| SPTO | 5578 | high | 6.1707 | 0.432 | 3.8% | 0.423 | 3.8% | 0.328 | 3.2% | 0.432 | |
| WEME | 2350 | high | 4.4368 | 0.495 | 5.4% | 0.5285 | 3.2% | 0.5714 | 3.1% | 0.5714 | |
| WBNU | 2895 | medium | 5.3122 | 0.973 | 1.6% | 0.849 | 4.5% | 0.638 | 3.3% | 0.973 | 4(10%) |
| WCSP | 3152 | medium | 4.7763 | 0.783 | 5.1% | 0.722 | 4.6% | 0.446 | 4.2% | 0.783 | |
| YBCH | 6627 | very high | 5.1501 | 0.604 | 4.4% | 0.469 | 4.4% | 0.366 | 4.0% | 0.604 | |

The results in Table 4 exhibit a bewildering span of performance. The TPRs for all three ensembles are quite impressive for some species (e.g., OCWA) while they all perform quite poorly for others (e.g., BEWR). Upon close examination of several of these results the following factors emerged as likely explanations for the performance variation.

1) Diversity of target species' sounds captured in the species dataset after preprocessing the input recordings (i.e., the post-preprocessed dataset)

2) Quantity of data in species post-preprocessed dataset

3) High-entropy species' sounds captured in the post-preprocessed dataset

4) The density of outliers in the species' post-preprocessed dataset

This study did not yield a quantitative assessment of the extent to which each of these factors influenced the outlier detection performance. Instead, we offer evidence in specific cases that support our conclusion that each of these factors is significant. First, we describe each of these factors in more detail and then we present the evidence in support of each factor using select species.

Factor #1: Diversity of target species' sounds captured in the post-preprocessed dataset

The diversity of sounds produced by the bird species included in this study is extensive both in terms of the number of different vocal sound types (e.g., song and number of call types), non-vocal sounds (e.g., drumming, bill snaps, wing whistles) and in terms of variability in songs from one individual of the species to the next. Some bird species have songs that are genetically programmed and so are innate and are produced with little variation from one individual to the next while other species learn their songs and have developed several distinct regional dialects. Since our datasets were not generally filtered for geographic regions, a variety of regional dialects might be captured. Still in other species individual birds develop a song repertoire that includes multiple song variations, referred to as songtypes (Pieplow, 2019). All these factors cause the diversity found across the species datasets included in this study to vary widely. In general, we found that the higher the diversity of target species sounds captured in the dataset the lower the accuracy of the outlier detection becomes.

Another related aspect of species sound diversity that was encountered in multiple datasets is that of low prevalence, but legitimate species sounds. One example of this issue is the sound that nestlings, fledglings and other juvenile birds make to stimulate adult birds to feed them. These calls are referred to here as begging calls and are generally distinct from calls made by adult birds of the same species. If the prevalence of such vocalizations is low relative to other target species sounds captured in the dataset, then they are often flagged as outliers. And since these are indeed legitimate targets species sounds detecting them as outliers results in a lower estimate of the outlier detection accuracy (i.e., false positives).

Species were grouped into four diversity categories based on the authors' reading the Birds of the World Research Review (Birds of the World, 2022) Sounds and Vocal Behavior section for each species. While this grouping is obviously subjective and, therefore, open to debate we believe that the concept that the diversity of sounds produced by different bird species can be ranked is indeed reasonable. The four categories used here are low, medium, high and very high. This categorization is included in Table 4 in the column labeled *diversity*.

Factor #2: Quantity of data in species post-preprocessed dataset.

It is widely accepted that successful training of deep learning models requires a minimum quantity of training data. The widely quoted rule of thumb is ten times the number of features but applied to our problem and considering each pixel of our 32x40 spectrograms as a feature, this level was not achieved for any of species included in the study. And while the number of post-preprocess spectrograms varied by more than a factor of 30, we did not find a strong linear correlation between the number of data and the accuracy of ensemble outlier detection. However, that is *not* to say that dataset size did not affect outlier detection as we certainly believe that it does, particularly at the lower end of the range of species dataset sizes included in this study. The quantity of post-preprocess data for each species is included in Table 4 in the number of clips column (i.e., *no. clips*).

Factor #3: High-entropy species' sounds captured in the post-preprocessed dataset

We observed that the false positive rate for some species' sound types was significantly elevated compared to other sound types in the same dataset and believe that this is due to the models having difficulty learning a compact representation for these sound types. We notice that issues seem to occur most frequently for percussive, noisy and buzzy sound types like drumming, snarls, harsh, raspy and other sounds with little tonal structure. Even when such sounds make up a substantial percentage of the target species sounds in the dataset they are sometimes flagged as outliers at high rates. Such a high false positive rate for even one of a species' sound types will lower the outlier detection accuracy and is particularly impactful if such a species sound makes up a substantial percentage of the sounds captured.

Spectrogram resolution plays a role here as well. If the relatively low 32x40 spectrogram resolution (adopted to keep the dimensionality manageable) reveals too little of the tonal structure of a species vocalization, then the spectrogram will also appear more noise-like. To quantify this affect we compute the Shannon entropy of each spectrogram per label derived from the original recording metadata from the Xeno-Canto repository. The overall mean of these entropy scores is included in Table 4 in the column labeled *entropy*.

Factor #4: The number of outliers in the species' post-preprocessed dataset

We found a few cases the performance of our outlier detection methods deviated from the trends observed in other datasets and could not be explained by the above factors. One last possibility that we investigated was that the number of true outliers present in the post-preprocessed dataset was just very low so the outlier detection algorithm which returns a fixed percentage of clips flagged as outliers returns a higher-than-normal number of false positives. Similarly, if the number of true outliers in the dataset is particularly high, then the outlier detection accuracy returned might also be higher. Of course, the only way to estimate the number of outliers in the overall dataset is to sample it which we decided was too expensive in terms of human effort to do in general. However, we sampled the *inlier class* of UOD ensembles for a select few species that exhibited peculiarly low and high TPR performance.

Figure 5 shows the best TPR performance of each ensemble type observed for each bird species included in the study. The species are grouped according to a rating of the diversity of the sounds that the species typically produces. Within each category, species are sorted from left to right in order of *decreasing* size of the species' post-preprocessed dataset.

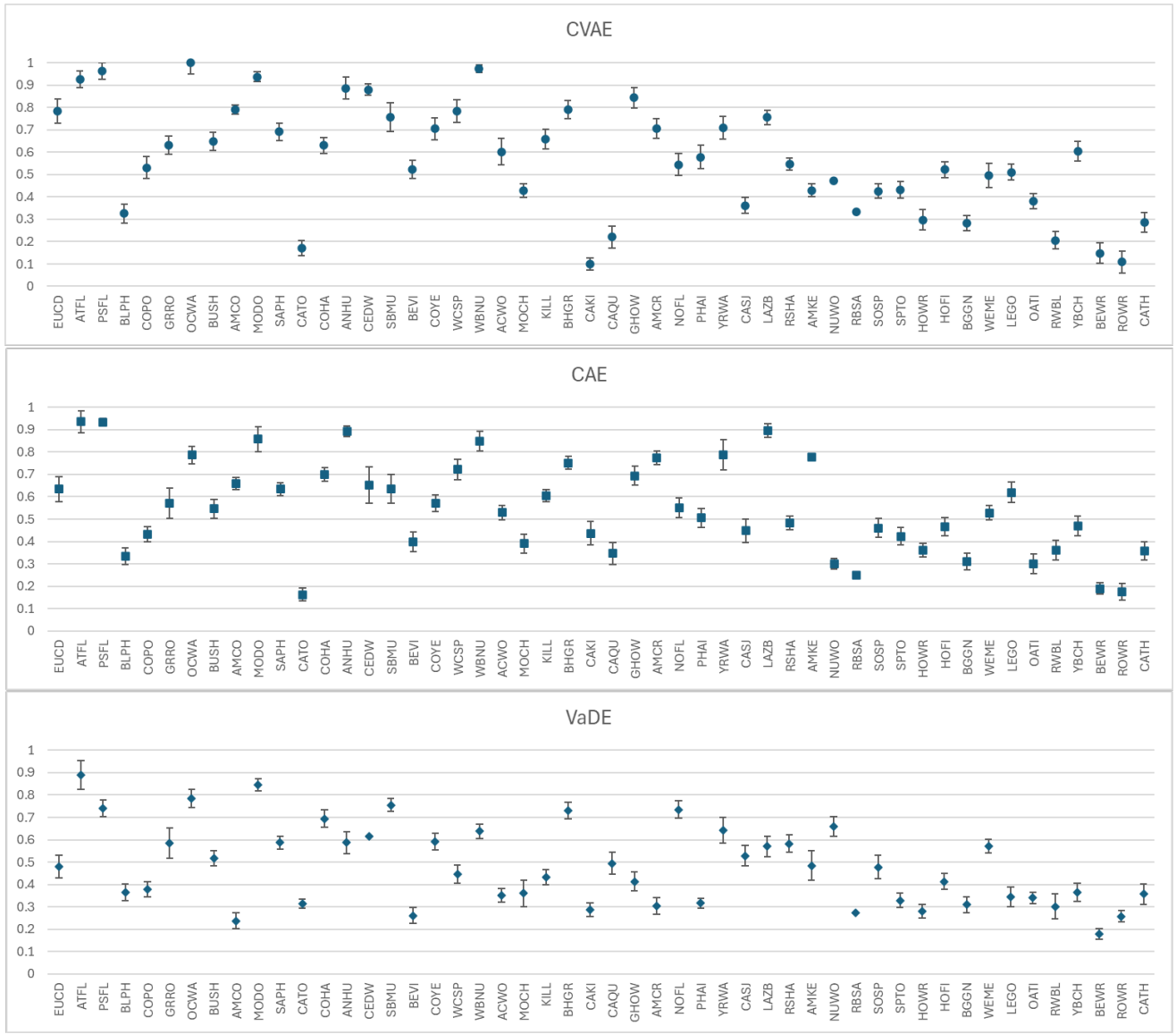

Figure 6. Best TPR performance of a) CVAE, b) CAE and c) VaDE ensemble type for each bird species with species grouped by low, medium, high and very high sound diversity from left to right.

## 11.1 Case for species sound diversity hindering UOD performance

The most apparent trend across the TPR performance plots in Figure 6 is that species that have lower sound diversity (i.e., species toward the left) exhibit better UOD accuracy on average than species with higher sound diversity toward the right. Figure 7 shows this more clearly by plotting the average TPR of each ensemble type for each of the sound diversity categories. Also included is the average TPR of the best ensemble type for each species in each category. This is understandable intuitively as the more sound types a species produces and the more complex each sound type becomes the more difficult it becomes for the models to learn a representation that captures the diversity.

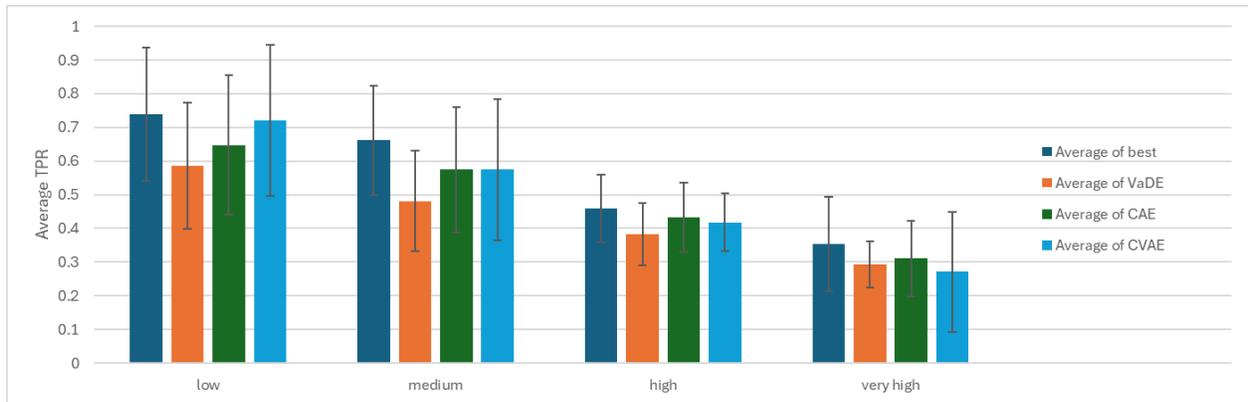

*Figure 7. Average TPR for each ensemble type across species in each diversity category. The average of the best ensemble's TPR is also plotted.*

Since the TPR performance trend is clear, species sound diversity is excluded from Table 4. The remaining factors are included for species for which the best TPR performance deviates significantly from the statistical mean in that diversity category in the column labeled *cause code*.

## 11.2 The case for low quantity of training data hindering UOD performance

As mentioned above, only a few of the bird species' datasets included in this study contained more than a few thousand spectrograms after preprocessing and some datasets only contained a few hundred. A few of these particularly small datasets were selected to investigate the impact of small training data size on performance. One big advantage of examining these smaller datasets is that the human effort required to quantify performance is significantly reduced.

The species with the lowest post-preprocess data size is Red-breasted Sapsucker (RBSA) with only 227 clips. All ensembles tested performed poorly with the best TPR obtained being 0.278, that is, only 27.8% of the outliers flagged by this best ensemble were verified to be true positives upon examination. And since the number of data was so small all the clips flagged by any ensemble were examined. Only six true outliers were found during this process and the best ensemble from each of the model types detected 5 of these 6 outliers. We also sampled the inlier class of the best ensemble to assess the false negative rate (FNR), something that was generally not done because of the human effort involved. This inlier class sampling revealed an estimated FNR of 14.6% ±5% (95% confidence). This says that there were plenty of other true positives in the dataset that these model ensembles did not detect.

The next lowest species dataset size is Nuttall's Woodpecker (NUWO) at 470 post-preprocessed clips. While still far below the rule-of-thumb quantity of training data this is more than twice the number of clips used in the model training for the RBSA. The best model ensemble trained on this NUWO dataset achieved a TPR of 0.659 which is a significant improvement over the best results for RBSA. Given that NUWO is another woodpecker species with similar sound diversity (including a similar percentage of high-entropy sound types) and similar FNR (13.7% ±5.3% 95% confidence), we interpret this improved performance for NUWO to indicate that the number of training samples for RBSA is just too low to adequately train any of the model types.

We conclude that there is a lower training dataset size threshold that needs to be exceeded in order to achieve reasonable UOD performance from these model ensembles. Do note however that the threshold appears to be significantly lower than the rule of thumb of ten times the number of features which indicates that these model ensemble methods can be effectively applied to smaller datasets.

## 11.3 The case for high entropy sound type(s) hindering UOD performance

This factor was recognized from outlier class sampling results. For some species it was noted that a high percentage of the clips that an ensemble included in the outlier class were in fact legitimate species sounds (i.e., false positives) but exhibited high entropy characteristics at the spectrogram resolution being used. Figure 8 provides an illustration of the issue using a VaDE model with only 2 latent space variables (z=2) overlayed with VaDE model (z=10) ensemble sampling results. Note that this z=2 model is used only for ease of graphical illustration and the sampling results are from an ensemble of z=10 VaDE models. Also note that only the ensemble's outlier class was sampled so, for example, the large and compact cluster around coordinates (-6, 6) contains latent space representations of the species' song clips which only a few were sampled because only a relative few song clips were in the outlier class. Contrast this to the much larger number of "whine" clips that were sampled. These are all false positives and their latent space representations are spread widely. Figure 9 shows four example spectrograms of the Bell's Vireo call type referred to in (Pieplow, 2019) as "whine". Note that not only does the number of sound bursts vary but also the spectrograms capture little tonal structure. The VaDE models failed to learn a compact latent space representation for this call type and our sampling of the outlier class for this ensemble showed that 79% of the false positives were categorized as "whine". This illustrates what we suspect is a significant factor for the relatively poor performance of our method on some bird species' datasets. The cause code column in Table 4 contains '3' for species which we suspect that this high entropy sound type has reduced TPR performance.

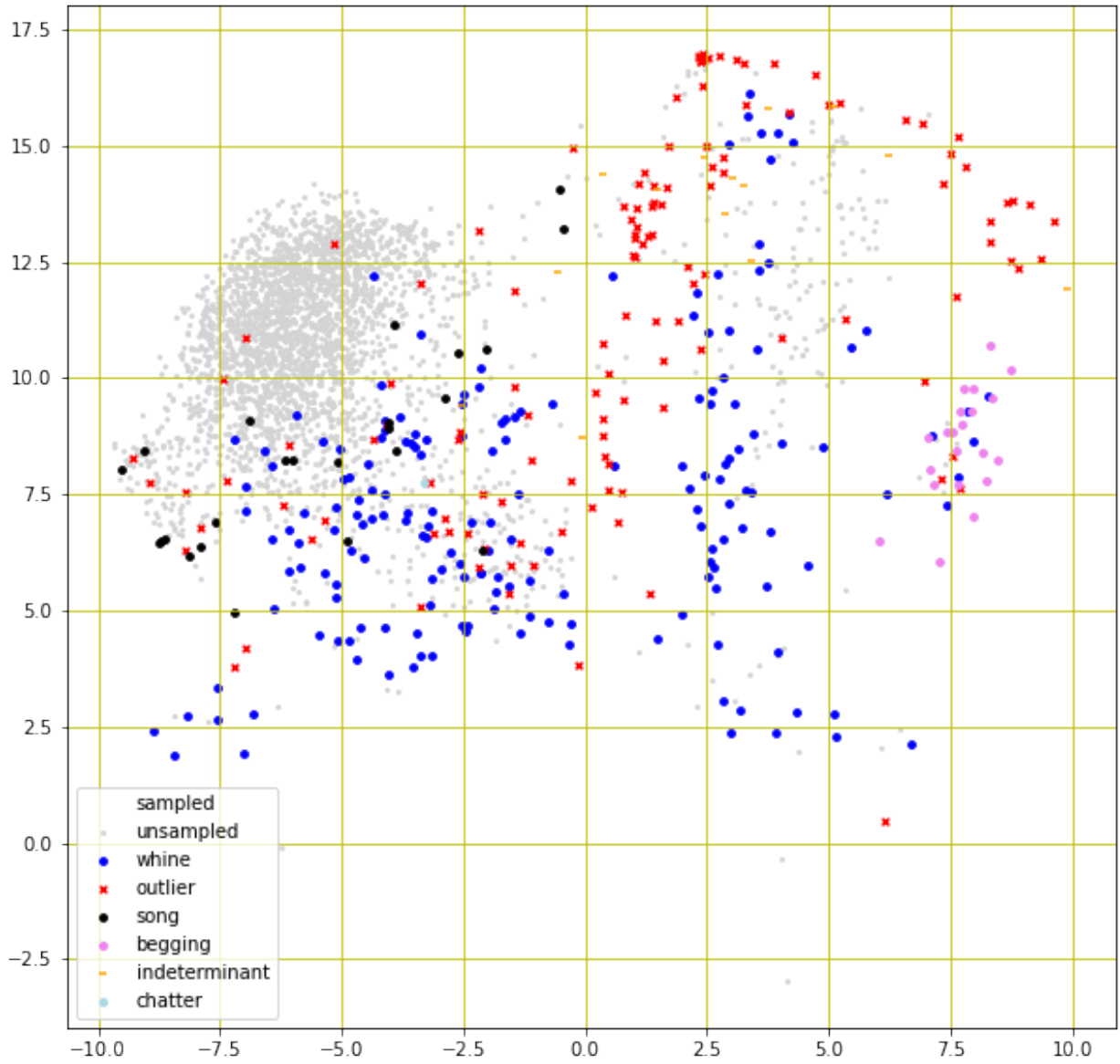

Figure 8. Scatter plot of one particular VaDE model latent space with 2 latent variables showing sampling results

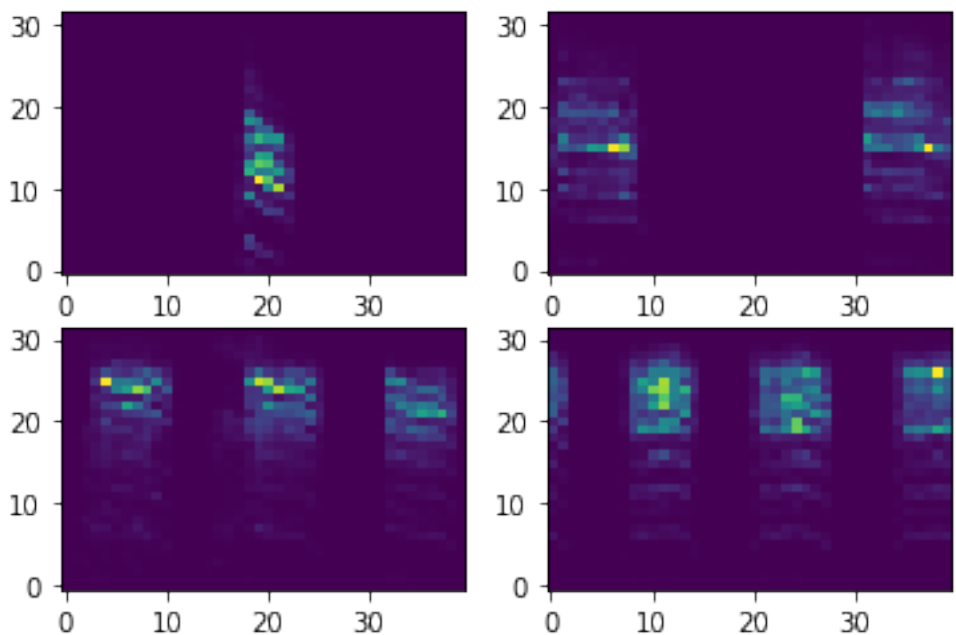

*Figure 9. Four examples of Bell's Vireo call type referred to as "whine"*

## 11.4 Case for density of outliers in the dataset affecting UOD performance

As mentioned above we found that the TPR performance results for a few species was significantly different from others in the same sound diversity category and could not be explained by either low clip count or high-entropy sound types. In our investigation of three such cases we resorted to sampling the inlier class of the model ensemble that achieved the best TPR performance to see if either a high or low density of outliers might be a cause. We selected Black Phoebe (BLPH) and California Towhee (CATO) because the TPR performance for both was significantly lower than that for other species in the low sound diversity category across all three ensemble types. We also sampled the inlier class for White-breasted Nuthatch because the TPR performance across the three ensemble types was unusually high than for other species in the medium sound diversity category. The margin of error for these sampling exercises was 5% with 95% confidence. In sampling the inlier class for BLPH we detected no false negatives and for CATO we found a 0.8% false negative rate (FNR). These results support our suspicion that low outlier density could be the cause for lower TPR performance. For WBNU we found a FNR of 10% which is quite high and since this was in the inlier class it indicates that the density of outliers in this dataset was significantly higher than the 10% outlier class limit imposed during UOD, thus supporting the hypothesis that a very high outlier density can result in better TPR performance.

## 12 Conclusions

This paper reports on our application of convolutional autoencoder and VaDE model ensembles to detect outliers in bird audio datasets derived from Xeno-Canto to improve on the bird species labeling. We found that the combination of the preprocessing steps and the dimensionality reduction achieved in the latent space representations of each of these model types significantly improved the ability to detect non-species sounds captured in these recordings. HAC applied to the latent space of the convolutional conventional autoencoder (CAE) and convolutional variational autoencoder (CVAE) was surprisingly effective relative to the more advanced and more complex Variational Deep Embedding (VaDE) models. Our results showed that both the CAE and CVAE ensembles outperformed the VaDE ensembles on average in each of the bird species sound diversity categories.

While our preprocessing steps involve only conventional signal processing the resulting cleaning should not be underestimated. Given the crowd sourced nature of the Xeno-Canto repository it should not be surprising that a large number of the recordings are simply not very good for our purpose. The preprocessing that we apply goes a long way to remove many subpar recordings and thus improve the performance of downstream UOD.

Simple majority-vote aggregation of multiple models significantly improves the accuracy of unsupervised outlier detection for all three model types.

We learned that a few factors influence the efficacy of the UOD techniques studied, primary among these are the diversity of bird species' sounds. We found that very small training dataset size and a high percentage of high-entropy sound types contained in the post-preprocessed dataset seemed to have a negative impact on performance. Additionally, we found evidence that unusually high density of outlier sounds in the post-preprocessed dataset can lead to better performance as measured by TPR and, conversely, that unusually low outlier density can lead to poorer performance.

We found that the simpler model ensembles consisting of CVAE coupled with HAC or CAE coupled with HAC performed, on average, better than the VaDE model ensembles on the dataset outlier detection tasks. Training VaDE models was more computationally expensive and problematic than training CVAE and CAE models on the bird sounds datasets and the need to specify more hyperparameters exacerbated this expense. Numerical stability issues while training VaDE models were particularly noticeable when the dataset size fall below ~1000 clips.

One should not underestimate the disadvantages of hyperparameters in unsupervised outlier detection. We were not able to find reliable model selection criteria and in the absence of such resorted to the laborious sampling of outlier classes to quantify performance. If the objective is a workflow that avoids the human effort of sampling, then we recommend CVAE for dimensionality reduction followed by the application of HAC to the latent space. However, we also believe that each of these model ensemble types were successful at detecting outliers in the bird sound datasets tested and, therefore, achieved the goal of reducing label noise.

# 13 References

(n.d.). Retrieved from https://xeno-canto.org/.

(2016), C. D. (2016). Tutorial on Variational Autoencoders. *http://arxiv.org/abs/1606.05908*.

Abhaya, A. &. (2022). An efficient method for autoencoder based outlier detection. *Expert Systems with Applications. 213. 118904. 10.1016/j.eswa.2022.118904.* .

Chalapathy, R. &. (2019). Deep Learning for Anomaly Detection: A Survey. *online at https://arxiv.org/abs/1901.03407*.

Denton, T. &. (2021). Improving Bird Classification with Unsupervised Sound Separation. *, Preprint at https://www.researchgate.net/publication/355141766_Improving_Bird_Classification_with_Unsupervised_Sound_Separation*.

Doersch, C. (. (2016). Tutorial on Variational Autoencoders. *http://arxiv.org/abs/1606.05908*.

Dong, C. &. (2018). The Feature Representation Ability of Variational AutoEncoder. *IEEE Third International Conference on Data Science in Cyberspace*.

Engel, D., Hüttenberger, L., & Hamann, B. (2012). A Survey of Dimension Reduction Methods for High-dimensional Data Analysis and Visualization. *OpenAccess Series in Informatics. 27. 135-149. 10.4230/OASIcs.VLUDS.2011.135.* .

*https://en.wikipedia.org/wiki/List_of_bird_genera*. (n.d.). Retrieved from Wikipedia.

*https://github.com/bcollins92078/bird_sounds_uod*. (n.d.). Retrieved from GitHub.

Jiang, Z. &. (2017). Variational Deep Embedding: An Unsupervised and Generative Approach to Clustering. *Twenty-Sixth International Joint Conference on Artificial Intelligence*.

Ma, M. &. (2023). The Need for Unsupervised Outlier Model Selection: A Review and Evaluation of Internal Evaluation Strategies. *ACM SIGKDD Explorations Newsletter. 25. 19-35. 10.1145/3606274.3606277.* .

Nielsen, F. (2016). 8. Hierarchical Clustering. In F. Neilsen, *Introduction to HPC with MPI for Data Science* (pp. 195–211). Springer.

Pieplow, N. (2019). *Peterson Field Guide to Bird Sounds.* New York, NY: Houghton Mifflin Harcourt Publishing Co.

S. M. Billerman, B. K. (2022). Retrieved from Birds of the World: https://birdsoftheworld.org/bow/home

Sun, I. D. (2022). A Survey of Ensemble Learning: Concepts, Algorithms, Applications, and Prospects. *IEEE Access, vol. 10*, pp. 99129-99149.

Wang, H. &. (2019). Progress in Outlier Detection Techniques: A Survey. IEEE Access. 7. 1-1. 10.1109/ACCESS.2019.2932769. *IEEE Access. 7. 1-1. 10.1109/ACCESS.2019.2932769*.

Zhang, W. &. (2019). Robust Class-Specific Autoencoder for Data Cleaning and Classification in the Presence of Label Noise. *. Neural Processing Letters. 50. 10.1007/s11063-018-9963-9.*